\documentclass[final]{cvpr}

\usepackage{times}
\usepackage{epsfig}
\usepackage{graphicx}
\usepackage{amsmath}
\usepackage{amssymb}
\usepackage{multirow}
\usepackage{multicol}
\usepackage{graphicx}
\usepackage{ifthen}
\usepackage{siunitx}

\newboolean{showImages}
\graphicspath{ {./images/} }
\usepackage[pagebackref=true,breaklinks=true,colorlinks,bookmarks=false]{hyperref}

\def\cvprPaperID{****} 
\def\confYear{CVPR 2021}
\newcommand{\norm}[1]{\left\lVert#1\right\rVert}

\begin{document}
\setboolean{showImages}{true}

\title{Generalizing GradCAM for Embedding Networks}

\author{Mudit Bachhawat\\
{\tt\small mudit5bachhawat@gmail.com}
}

\maketitle

\begin{abstract}
   Visualizing CNN is an important part in building trust and explaining model's prediction. Methods like CAM\cite{cam_arxiv} and GradCAM\cite{gradcam_arxiv} have been really successful in localizing area of the image responsible for the output but are only limited to classification models. In this paper, we present a new method EmbeddingCAM, which generalizes the GradCAM for embedding networks. We show that for classification networks, EmbeddingCAM reduces to GradCAM. We show the effectiveness of our method on CUB-200-2011 dataset and also present quantitative and qualitative analysis on the dataset.
\end{abstract}

\ifthenelse{\boolean{showImages}}{

\begin{figure*}[t] 
\centering
\includegraphics[width=\textwidth]{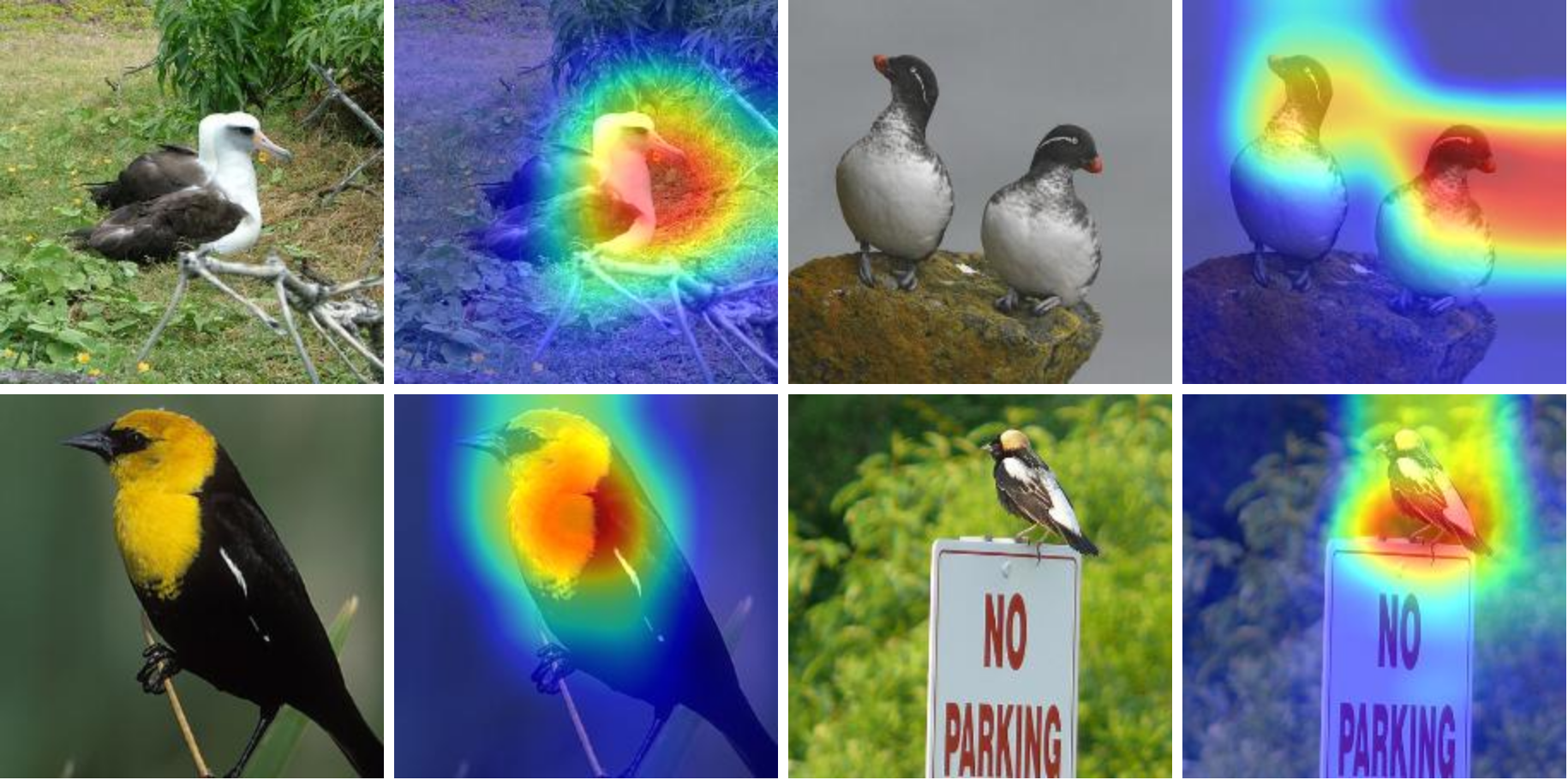}
\caption{Sample results generated using mean proxy method}
 \label{result_diagram}
\end{figure*}

}{}

\section{Introduction}
Convolutional neural networks are widely used for image classification, image captioning, face recognition, representation learning, open set classification, visual recommender system. Generally neural networks are considered as black box and the output generated by the model is always in question. Thus, visualization tools are very important to build confidence and trust in the neural networks. ~~Neural networks generally trained end-to-end makes it hard to debug and explain the output generated.~~ For example: a product recommender system must know which parts of the image the model is looking at to generate the recommendations. A bird classification model must look at the bird area to generate the predictions instead to any background noise to generate the prediction. This kinds of explanations helps in building trust in the neural networks. 

Interpobaility methods like 
\cite{cam_arxiv, gradcam_arxiv, gradcampp_arxiv, ZeilerF13}
are widely used for classification and tasks with classification branches. But these methods fail to generalize on embedding network. Embedding networks outputs embeddings instead of discrete class scores. Embeddings, generally in context of neural networks, are learned vector of continuous variables, which generally represents a discrete class. Embeddings are useful as they captures low dimensional rich representation of otherwise high dimentional data. For example the task of capturing of the body pose can be represented by a embedding \cite{pose_embeddings} rather than representing it in discrete classes (as body pose is a continuous measure and the scale of unique poses can be huge). 

Few methods have tried to visualize embedding networks. Methods like \cite{DBLP:journals/corr/abs-1811-07487} have tried to generate visualization heatmaps on embedding network, but have trained a classifier on top of embeddings. This type of classification head is not generally available in embedding networks. Another method by \cite{DBLP:journals/corr/abs-1901-00536} successfully generates heatmap to focus areas in images, which are responsible for similarity between pair of images. This method requires a pair of images, which is not a requirement of a embedding network. Chen \textit{et al.} \cite{adaping_Chen_2020_WACV} provides a method to generate localized heatmap for single image, but requires sampling of multiple triplets for loss. Additionally, the images needs to be selected properly as their embeddings should violate the maximum condition for triplet loss to flow. Our method doesn't require any such sampling and works with single image. 

Our method EmbeddingCAM generalizes GradCAM and is analogous to GradCAM for classification type networks. For classification, GradCAM generates class discriminative map, focusing regions of images which are responsible in generating the score for that class. The generalization of GradCAM is hard for embedding network as embeddings cannot be directly used for back-propagating without the class scores. We solve this problem by using proxies as class representative and following a similar method as GradCAM. Under special conditions, i.e. when embeddings are substituted with one-hot encoded vectors, we show that EmbeddingCAM reduces to GradCAM. In a way, embedding is a generalization of classification vector, just trained on different loss. Our method is well represented in figure \ref{main_process_diagram}. 

In summary our work has following major contribution

\begin{itemize}
    \item We provide a method to generate a GradCAM type heatmap to highlight regions of images, which the network is focusing at, to generate the embedding
    \item We provide a generic method of EmbeddingCAM, which reduces to GradCAM when adjusted for classification task
    \item We conduct experiments open set classification method, to show the working of our method, by generating heatmaps on CUB-200-2011 dataset
    \item We provide qualitative and quantitative metrics of our method on CUB-200-2011 dataset. We also show the comparison with previous methods
\end{itemize}


\section{Related Works}

In this section, we explore related works on visual embedding networks and interopability of neural networks. 

\textbf{Visual Embedding Networks}: Visual embedding networks outputs a fixed sized embedding for a given input image. The network can be then trained using loss which uses relative distance between these embedding. Additionally, to add boundedness into the network output, $L_2$ normalization is generally added and cosine similarity is widely used. Such models are trained using Triplet Loss or Siamese Loss. Finding new losses and efficiently training embedding networks is an active area of research \cite{proxynca}\cite{proxyncapp}\cite{Kim_2020_CVPR}. Once the network is trained, our method can work on any network, trained on any loss.

Most of the methods to train these networks is either a \textit{pair-based} method or a \textit{proxy-based} method. Pair-based methods minimizes the relative distance of the images in the embedding metric space. For example, triplet loss \cite {triplet1_DBLP:journals/corr/WuMSK17, triplet2_DBLP:journals/corr/SchroffKP15} samples three images, two from same class and one from different class, and minimises the relative distance of samples from same class to samples from different classes. Another example, contrastive loss \cite{contrastive_1640964, contrastive_chopra2005learning}, which aims to minimize the distance between a pair of data if their class labels are identical and to separate them otherwise.  

Pair-based methods requires sampling of multiple images during training time. This increases the training complexity and thus creates a need for efficient mining of hard and semi hard negatives. Proxy-based methods  \cite{Kim_2020_CVPR} \cite{proxynca} \cite{proxyncapp} uses proxy embedding to optimize the sampling complexity and improve the training process. These methods use proxies to compute the loss between proxy and embedding, thus preventing use of any other embeddings.

Common methods to train a these models is using triplet loss \cite{triplet1_DBLP:journals/corr/WuMSK17, triplet2_DBLP:journals/corr/SchroffKP15, Koch2015SiameseNN}. Generally these methods use relative distance between images to supervise the model. Common use cases for these systems includes open set classification, visual search engine, etc. 

\textbf{Explaining Deep Networks}: CAM\cite{cam_arxiv} and Grad-CAM\cite{gradcam_arxiv} are widely used methods of visualizing classification networks. These methods highlights important areas in image which are responsible for generating the class scores for a specific class. Limitation: These methods are only limited to classification type networks, and not apply for embedding type networks.

GradCAM \cite{gradcam_arxiv} and related methods \cite{gradcampp_arxiv} are widely used for visualizing classification type networks. GradCAM backpropagates the gradient of class score ($y_c$) to the last convolutional layer and reduces those gradient weights with last convolutional layer output to generate final heatmap. For simple networks (i.e. Conv -> Avg Pool -> FC), \cite{gradcam_arxiv} has shown GradCAM reduces to CAM \cite{cam_arxiv}, which uses fully connected layer weights to generate heatmap.

\textbf{Visualizing Embedding Networks}: Chen et al. \cite{adaping_Chen_2020_WACV} proposed a method to generate Grad-CAM type heatmap for embedding network. Their method requires multiple training examples to first compute gradient weights for training samples. Then, during testing phase, they find the nearest training sample to the testing sample, and use gradient weight of the training sample to generate the heatmap. This method require multiple images and requires to build an index to generate heatmap for any arbitrary image.

\ifthenelse{\boolean{showImages}}{

\begin{figure*}[t] \label{main_process_diagram}
\centering
\includegraphics[width=\textwidth]{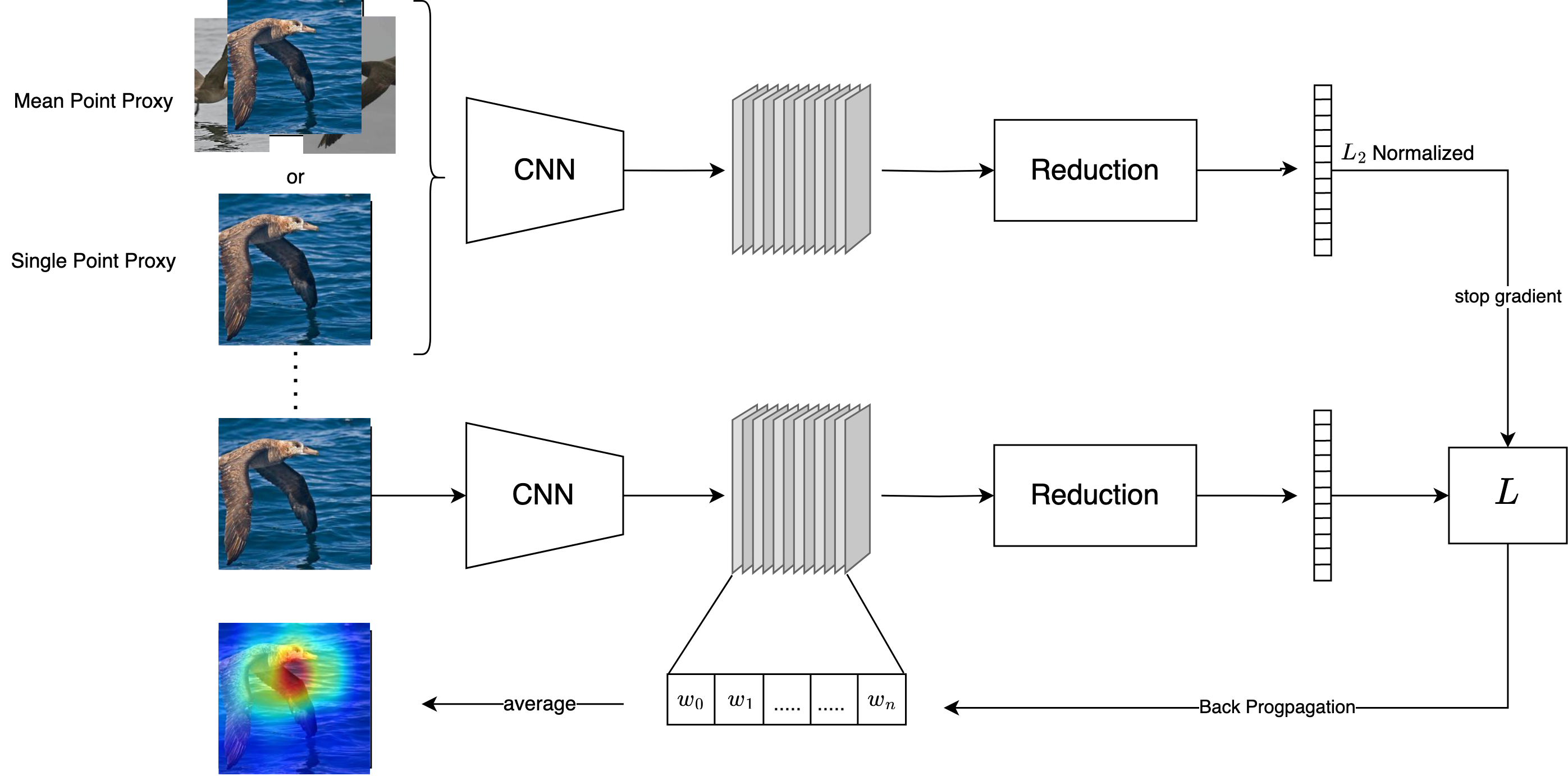}
\caption{Diagram showing our method for generating heatmap from embedding models}
\end{figure*}

}{}

\section{Method}

In this section, we go through the method of generating a Grad-CAM style heatmap to highlight important parts in image. A high level visualization of our method can be found in figure \ref{main_process_diagram}. We use proxy as class embedding to compute a custom loss[\ref{new_loss}]. Then, similar to GradCAM we use back-propagated weights to generate final heatmap. Since the loss depends the proxy used, we explore couple of method to generate the proxy. The proxy can be imagined as a ground truth vector, i.e. a vector for the concept we are trying to visualize. More on this in section \ref{sec:proxies}. 

\subsection{Preliminary}
This section covers the preliminary topics on which we build our method.

\subsubsection{Visual Metric Learning}
Visual metric learning is the task of learning a function $f: I \rightarrow S$ which maps an image $I$ to a semantic space $S = \mathbb{R}^d$. This kind of formulation allows to capture rich representation in the data. The embedding of semantically similar images should be closer and embeddings of semantically dissimilar images should be farther in the metric space $\mathbb{R}^d$. Generally the function $f$ is replaced with a convolutional neural network which outputs an embedding in $\mathbb{R}^d$. Euclidean distance is used on the $L_2$ normalized embedding to measure distance between points. 

\begin{equation}
    y = \mathrm{CNN}(I)
\end{equation}

\begin{equation}
    \hat{y} = \frac{y}{\norm{y}}
\end{equation}

where $\norm{.}$ is $L_2$ norm and $\hat{y} \in \mathbb{R}^d$

\subsubsection{GradCAM}
GradCAM \cite{gradcam_arxiv} generates a class discriminative map (heatmap) for classification type network. It uses gradient of the class score, to calculate the area of image responsible the output score. It has been used to generate class local heatmaps for various tasks like image classification \cite{gradcam_arxiv, gradcam_appl_DBLP:journals/corr/abs-1901-07031, gradcam_appl_9204258}, visual question answering \cite{gradcam_appl_DBLP:journals/corr/abs-1908-06306} and image captioning \cite{gradcam_appl_DBLP:journals/corr/abs-2001-01037}.

Mathematically, Let $Y_c$ be the class score (pre softmax) of the discriminative class. We then compute the gradients of $Y_c$ with respect to last convolutional layer of the network $A^k_{i,j}$ (generally before global average pooling)

\begin{equation} \label{gradient_gradcam}
g_{c}\left(A^{k}_{i,j}\right)=\frac{\partial y^{c}}{\partial A^{k}_{i,j}}
\end{equation}

where $(i,j)$ are the spatial indexes of the tensor and Z is the total number of spatial pixels in each channel. We then reduce the gradient tensor across spatial dimensions, to generate channel weights. 

\begin{equation} \label{reduction_eqn}
\alpha_{k}^{c} = \frac{1}{Z} \sum_{i} \sum_{j} \frac{\partial y^{c}}{\partial A_{i, j}^{k}}
\end{equation}

We finally reduce the last convolutional layer output using the channel weights obtained from equation  \ref{reduction_eqn}.

\begin{equation} \label{gradcam_final}
H_{\mathrm{Grad}-\mathrm{CAM}}^{c}=\operatorname{ReLU} \left(\sum_{k} \alpha_{k}^{c} A^{k}\right)
\end{equation}

\textbf{Computations for simple networks}: For simple networks, i.e. final convolutional layer followed by global average pooling, followed by dense layer, followed by softmax layer for classification. The equation \ref{reduction_eqn} reduces to

\begin{equation} \label{gradcam_cam_relation}
    \alpha_{k}^{c} =  \frac{1}{Z} w_{k, c}
\end{equation}

Where $w \in \mathbb{R}^{K \times C} $ is the kernel of dense layer, with number of channels in last convolutional layer K and number of classes C. Ignoring the $1/Z$ factor, which will be normalized during the visualization, the $\alpha_{k}^c$ is identical to $w_{k,c}$. The proof of equation \ref{gradcam_cam_relation} can be found in \cite{gradcam_arxiv}. The channel weights can be arranged in vector as 
\begin{align}
    \alpha^c &\propto (w_{1, c}, w_{2, c}, ..., w_{K, c})^T \\
    \label{alpha_init_relation}
             &\propto w \cdot o^c
\end{align}

Where $o^c$ is one-hot encoded vector for class $c$. In the following chapters, we will show how $o^c$ can be visualized as proxy vector of EmbeddingCAM. 

\subsection{Generalizing GradCAM for Metric Learning}
As mentioned in \cite{adaping_Chen_2020_WACV}, applying GradCAM to metric learning have two major challenges. First the is no class labels ($Y_c$). In GradCAM, the gradients for convolution output is calculated with respect to the class score. Since most embedding networks are trained using relative distances between embeddings, there is no notion of class score. Secondly, because of there is no notion of class, it becomes extremely hard to compute the gradient with one sample. You will need to take multiple samples to compute the gradient.

Our method solves both the problem by using proxies as class representative. Formally, let $p_c \in \mathbb{R}^d $  be the proxy of class c with dimension d. We define the process of calculating proxies in section \ref{sec:proxies}.  We further define following as the loss function, for calculating the gradients.

\begin{equation} \label{new_loss}
\mathcal{L}_c = y \cdot p_c
\end{equation}

where '$\cdot$' represents dot product and  $y \in \mathbb{R}^d$ is unnormailzed CNN output. This is analogous to using pre-softmax class score $Y_c$ in GradCAM. Finally, we use $\mathcal{L}_c$ to calculate gradients similar to \ref{gradient_gradcam}.

\begin{equation}
g_{c}\left(A^{k}_{i,j}\right)=\frac{\partial \mathcal{L}_{c}}{\partial A^{k}_{i,j}}
\end{equation}
We use equation \ref{reduction_eqn} and \ref{gradcam_final} to calculate the final heatmap.

\subsubsection{Calculating Proxies} \label{sec:proxies}
We explore two schemes for calculating proxy of class C.

\textbf{Normalized Mean Proxy}: For any class c with images ${i_1, i_2, ...i_n}$, we calculate average proxy for class c as
\begin{equation}
p_c = \frac{\frac{1}{n}\sum_{j=i}^n \hat{y}_j}{\norm{\frac{1}{n}\sum_{j=i}^n \hat{y}_j}}
\end{equation} where 
\begin{equation}
\hat{y_j} = \frac{CNN(i_j)}{\norm{CNN(i_j)}}
\end{equation}\textbf{Single Point Proxy}: Single Point Proxy is a special case of normalized mean proxy where we use only one point i.e. the image embedding itself as class representative. This method solves the problem of sampling multiple images to generate a heatmap.

\subsubsection{Intuition}
In GradCAM, we calculate the gradient of class score $y_c$ w.r.t the final convolutional output. The gradients in GradCAM highlights the area, when magnified, increases the class score. Thus $\alpha_k^c$ represents direction of each channel in final convolutional output, which can lead to increment in the class score $y_c$. This direction can be also visualised as weights for each channel in the generation of final heatmap. In this process the gradients create class discriminative activation map for the class.  

Similar to this, the equation \ref{new_loss} represents the agreement between the model output and the class proxy. If we think of class proxy as the ground truth vector of the concept for which we are trying generate visualization, the equation \ref{new_loss} is analogous to class score of the Grad-CAM. 

The equation \ref{new_loss} can be written as $|y|cos\theta$, where $\theta$ is the angle between model output vector and proxy vector. Thus the region highlighted by the gradient of this loss is responsible for increasing the model output confidence ($|y|$) and agreement between proxy and model output vector. Additionally, if y is an output of classification network with number of classes $C$ and $p_c$ is an one hot encoded vector for class $c$, the equation \ref{new_loss} reduces to class score of class $c$. This reduces EmbeddingCAM to GradCAM.

\subsection{Reduction for Simple Networks}
For simple image classification type networks (networks ending with convolutional layer, followed by average pooling, fully connected and L2 normalization layer to generate final embedding), we simplify the network calculations and draw similarity between GradCAM and EmbeddingCAM. Let $w \in \mathbb{R}^{K \times C} $ be the kernel of last fully connected layer. Continuing from equation \ref{new_loss}:

\begin{equation}
    \frac{\partial L_c}{\partial y} = p_c
\end{equation}

Back-propagating the loss through fully connected layer and average pooling layer, we get following as gradients:

\begin{equation}
    \frac{\partial \mathcal{L}_{c}}{\partial A^{k}_{i,j}} = \frac{1}{Z} \sum_{x=1}^{C} w_{k,x} * p_{c,x} 
\end{equation}
Since the above gradient is independent of $i$ and $j$, $\alpha_k^c$ can be reduced to \ref{final_alpha_reduction}
\begin{align} 
    \alpha_k^c & = \frac{1}{Z} \sum_{i} \sum_{j} \frac{\partial  \mathcal{L}_{c}}{\partial A_{i, j}^{k}} \\
    \label{final_alpha_reduction}
               & = \frac{1}{Z} \sum_{x=1}^{C} w_{k,x} * p_{c,x}
\end{align}

Finally, we vectorise the equation \ref{final_alpha_reduction} to generate \ref{vectorised_reduction}.

\begin{equation} \label{vectorised_reduction}
    \alpha^c = \frac{1}{Z} w \cdot p_c
\end{equation}

\begin{equation} \label{alpha_similar_relation}
    \alpha^c \propto w \cdot p_c
\end{equation}

We again ignore the $1/Z$ factor in equation \ref{vectorised_reduction}. This factor will get normalized during  heatmap generation. Please note that equation \ref{alpha_similar_relation} is similar to \ref{alpha_init_relation}.

\subsection{Comparison with other methods}

Method \cite{adaping_Chen_2020_WACV} requires selective sampling of multiple triplets for a single image to generate the heatmap. Additionally, this methods uses training data weights during inference phase. This, the method is not generic and cannot work with just pretrained model. Our method can work with just pretrained model and can work with single image (when using single point proxy scheme).



\begin{table*}[t]
\begin{center}
\begin{tabular}{|c|c|c|c|c|c|}
\hline
\multicolumn{6}{|c|}{Visual Attention Accuracy}                                                                                                                   \\ \hline
\multicolumn{1}{|l|}{}                           & \multicolumn{4}{c|}{Mean Heatmap Ratio}           & \multicolumn{1}{l|}{\multirow{2}{*}{Accuracy (IoU @ 0.5)}} \\ \cline{1-5}
                                                 &      & Uniform & All Channels  & Top-50  Channels & \multicolumn{1}{l|}{}                                      \\ \hline
\multirow{2}{*}{Chen et al. \cite{adaping_Chen_2020_WACV}}    & BBox & {0.543}   & {0.643 $\pm$  0.265} & \textbf{0.760 $\pm$ 0.152}    & \multirow{2}{*}{\textbf{0.797}}                                     \\ \cline{2-5}
                                                 & Mask & 0.275   & 0.416 $\pm$ 0.231 & \textbf{0.534 $\pm$ 0.146}    &                                                            \\ \hline
\multirow{2}{*}{Our Method (Single Point Proxy)} & BBox & 0.522   & \textbf{0.776 $\pm$ 0.004}        & 0.757 $\pm$ 0.003                & \multirow{2}{*}{{0.776}}                                     \\ \cline{2-5}
                                                 & Mask & 0.274       & \textbf{0.543 $\pm$ 0.004}         & 0.520 $\pm$ 0.004                &                                                            \\ \hline
\multirow{2}{*}{Our Method (Average Proxy)}      & BBox & 0.522       & \textbf{0.789 $\pm$ 0.005}            & 0.764 $\pm$ 0.004               & \multirow{2}{*}{{0.768}}                                        \\ \cline{2-5}
                                                 & Mask & 0.274      & \textbf{0.551 $\pm$ 0.006}            & 0.524 $\pm$ 0.005               &                                                            \\ \hline
\end{tabular}
\end{center}
\caption{Comparison of our method to \cite{adaping_Chen_2020_WACV}. The accuracy reported is weakly-supervised localization accuracy with IoU@0.5. We train 8 models with different initializations to provide standard deviation of the score. }
\label{results_table}
\end{table*}


\section{Experiments}

\subsection{Dataset: CUB200-2000}


We use CUB200-2011 dataset for the evaluation of our method. The dataset contains 11.8K images of birds across 200 classes. This dataset is commonly used for fine grained classification, open set classification, and visual explanations. We trained a ResNet-50 embedding model following the method details by \cite{wu2017sampling}, which outputs embedding of size 128. We use the output of last convolutional layer (before average pooling layer) of size $7\times7\times2048$, for back-propagating the loss and generating the heatmap. The model is trained on first 100 classes and evaluated on next 100 classes. 

\textbf{Evaluation of the method:} Our method takes input a pretrained model which is used to generates heatmaps on the complete dataset. The heatmaps generated are evaluated using the metrics defined below. Please note that bounding boxes and segmentation maps are only used during the evaluation phase, and are not used during the training of the model.

\textbf{Metrics:} We use the mean heatmap ratio metric to evaluate our method. Mean heatmap ratio is the ratio of the heatmap score inside the bounding box (or segmentation map). We calculate this by calculating ratio of heatmap inside the bounding box (or segmentation map) and then taking mean over all images. We also present a uniform baseline, which calculates this metric on a constant heatmap (heatmap that have constant non zero activation in its spatial dimensions). For a heatmap which has activation on bird area, the heatmap ratio will be one for both bounding box and segmentation map. 

We also report weakly supervised localization accuracy as reported in \cite{adaping_Chen_2020_WACV}. This metric requires computation of bounding box from the heatmap and calculates intersection over union (IoU @ 0.5) accuracy with the ground truth bounding boxes. This metric measures the bounding box prediction with weakly supervised training of the model. This metrics is widely used for visual interpretation tasks \cite{adaping_Chen_2020_WACV}\cite{wsol_example}. 

To generate the bounding box from heatmap, we first binarize the heatmap with a threshold $t$. We then find the maximum connected component in binarized heatmap. Finally we take the smallest bounding box that completely encloses the largest connected component. 



\subsubsection{Results}

We compare our method with \cite{adaping_Chen_2020_WACV}, which also reports mean heatmap ratio and weakly supervised localization accuracy. We have attempted our best to keep the environment same while generating the results. Please note that bounding boxes and segmentation maps are only used during evaluation phase and not during training of the model.


\textbf{Mean Heatmap Ratio}: We calculate mean heatmap ratio for all classes using bounding box and segmentation map. The report our metrics in table \ref{results_table}. Our method both single point proxy and average point proxy), both performs competitive to \cite{adaping_Chen_2020_WACV} without the need for sampling multiple triplets and test-time indexing. Our method does not require sampling and thus mean heatmap ratios are stable (doesn't have variance) compared to \cite{adaping_Chen_2020_WACV}. We additionally train 8 models with different random initialisation to test the stability of our model. We report the standard deviation of mean heatmap ratio in the table. Our single point proxy achieves mean heatmap ratio of 0.776 and 0.543 for bounding box and segmentation map respectively. We observe the standard deviation of our methods to be less than 1\%, demonstrating stability of our approach. 

\textbf{Weakly Supervised Localization Accuracy}: We report weakly supervised localization accuracy in table \ref{results_table} using the threshold $t=0.2$. We have found the accuracy to be stable ($> 0.6$) for all values of $t \in [0.05, 0.4]$. We calculate all our metrics


\textbf{Qualitative Analysis:} Figure \ref{result_diagram} presents sample images with their heatmaps generated by our method. We observe that most of our heatmaps are generated on relevant parts of birds which uniquely identifies the species.  

\section{Conclusion}

In this paper, we present a novel approach to explain the prediction of any metric learning model. A wide variety \cite{cam_arxiv}\cite{gradcam_arxiv}\cite{gradcampp_arxiv} of methods have been developed to explain the prediction of a classification system. But, as the application of metric learning are increasing there are no simple and effective method to explain the prediction of the system.

In this paper, we present a novel method to generate GradCAM style heatmap from just a pretrained metric learning model. We do this without the use of multiple images and selective sampling of multiple images. We are able to achieve this by using proxies as class representative and and defining a custom loss (which can be used for backpropagation, similar to GradCAM). Finally, we show the working of our method in the results section. We show that we have achieved comparable mean heatmap ratio and weakly supervised localization accuracy in comparison to \cite{adaping_Chen_2020_WACV} on CUB200-2011 dataset. We also present some qualitative analysis of the results generated. 

The recent advancements in the field of AI has greatly enhanced the accuracy of the tasks. But, deep learning method primarily being a black box system has caused havoc when deployed in production. To avoid such incidents, it is extremely important for ML models to be a white box rather than black box and developing an explainable AI system plays a huge role in this. We encourage researchers to apply this method on other embedding networks share the effectiveness of the approach. 

{\small
\bibliographystyle{ieee_fullname}
\bibliography{egbib}
}

\end{document}